# Exploring Fact Memorization and Style Imitation in LLMs Using QLoRA: An Experimental Study and Quality Assessment Methods

**Authors**:
Eugene Vyborov, Oleksiy Osypenko, Serge Sotnyk - ability.ai

## Abstract

There are various methods for adapting LLMs to different domains. The most common methods are prompting, finetuning, and RAG. In this work, we explore the possibility of adapting a model using one of the PEFT methods - QLoRA. The experiment aims to simulate human responses based on their interviews. The simulation quality is assessed by comparing the quality of the style and the quality of the generated facts.

## Introduction

Large language models (LLMs) such as ChatGPT, Mistral, and LLaMA demonstrate remarkable capabilities in solving a wide range of tasks across various domains, including answering questions [1], solving mathematical problems [2], generating code [3, 4], and much more. However, adapting a model for a specific task can significantly improve its performance, although this may come at the cost of the model's generality.

The simplest way to adapt is by writing specific commands - prompts that instruct the model on what it should do [6]. This method is advantageous because we can use almost any LLM - both public ones like Mistral v0.3 7b or LLaMA 3 8b (in both cases, instruct versions must be used) and private ones available through an API. However, this solution has drawbacks: if we have a lot of information and facts that need to be used, they may not fit in the context window of the selected model. Moreover, this method makes it challenging to regulate the presentation style and set some complex entities like the "perspective" we want to copy from a subject matter expert.

To mitigate issues caused by prompt volumes, the Retrieval-Augmented Generation (RAG) method is often used [5]. In this case, additional context is selected for each model request, usually through vector databases, but it can also be a regular search engine or some other specific mechanism. This context may include facts, rules for decision-making, and more. This method significantly expands prompting capabilities, giving the model and agents based on it an analog of associative memory. However, it does not entirely eliminate other complexities - prompts need to be written carefully, analyzing their impact on the result. Prompts (especially complex ones) mixed with RAG context consume the context that could be used for data processing, increasing the cost of each request to the LLM. Despite this, using prompting and RAG is a good first step for many projects, often solving quite complex tasks with acceptable quality at a reasonable cost (including product development cost).

Of course, it would be ideal if the necessary knowledge and skills were already embedded in the model solving the task, eliminating the need to additionally instruct it or add fragments of knowledge and rules. For instance, we want it to respond in the style of a particular expert. To achieve this, we can try finetuning and aligning the model to the task. In this case, the knowledge about the task and response style will be contained in the model's weights, eliminating the need to add corresponding information in the prompt and context.

This approach has its drawbacks - preparing a training dataset, training the model, and ensuring that after training, there is no catastrophic forgetting of the information needed for the task. The balance of these pros and cons should be considered when making architectural decisions on how the LLM is used in the project.

In this article, we consider one of the finetuning options - Parameter-Efficient Fine-Tuning (PEFT) [7], specifically one of its widely used representatives - QLoRA [8] (a fine-tuning method that combines Quantization and Low-Rank Adapters - LoRA).

Another important aspect considered in the article is the quality assessment of fact memorization.

Experiment Essence:

- We aim to train a model to respond in the style of an expert. Besides the style, it is important that the model knows the facts on which the expert bases their decisions.
- As a knowledge model of the expert, we will use a set of interviews with a well-known media personality to make it easier to find the necessary data in open access. In our study, we used interviews with Elon Musk.
- An interesting question is the volume of required data.
- We will develop a way to assess the correspondence of style and especially the factual content in the model's responses.

The experiment codes, datasets, and report data can be found on GitHub:
https://github.com/Abilityai/research-expert-mimic

# Training

## Dataset Preparation

The dataset is prepared from publicly available interviews. Interviews were parsed, and sequences were formed corresponding to the format supported by the transformers framework for chat message generation (https://huggingface.co/docs/transformers/en/chat_templating).

For instance:

```
chat = [
   {"role": "user", "content": "Hello, how are you?"},
   {"role": "assistant", "content": "I'm doing great. How can I help you today?"},
   {"role": "user", "content": "I'd like to show off how chat templating works!"},
]
```

The person of interest was marked as the assistant, while all others (more than one person might participate in the interview, sometimes questions come from the audience) were marked as the user.

Various interview artifacts were removed, and the order of sentences was adjusted to the correct alternation of user and assistant roles for most models (consecutive messages from one role were merged).

Initially, the dataset volume was about 400 KB of plain text (5-6 hours of interviews). Preliminary experiments showed that this amount was insufficient for quality model training, so the data volume was increased to 2 MB of plain text (slightly more than a day of interviews).

For style verification, files starting with an underscore were used. For fact-checking, files containing the substring "2023" in their name were used as they were the freshest interviews in the dataset.

Final training was conducted on fragments cut from interviews, with a length of N sub-dialogs (two consecutive user + assistant messages). N varied from 2 to 8 in preliminary experiments. Such fragments were formed using a sliding window that slid along user messages at the beginning and assistant messages at the end.

Then the beginning of such a fragment was trimmed to a random length from 1 to N.

The dataset used can be found at the following link:
https://github.com/Abilityai/research-expert-mimic/tree/main/data/chat.

## Fact Q/A dataset

It is difficult to verify the possibility of fact memorization from the training dataset on the test dataset because those facts might simply not be there. Therefore, investigating fact memorization required an additional dataset called the fact q/a dataset. It was prepared using the freshest interviews from the main dataset - three interviews from 2023. These interviews were processed using ChatGPT 4o, forming the following columns:
- **fact**: Description of the fact found in the text.
- **scr**: Original fragment from which the fact was taken. This column is for ease of verification.
- **question**: Question that allows understanding if the person asked this question knows the fact.
- **answer**: Fact recorded as a direct answer to a question about this fact - since we mimic a person in the form of a chat, comparing with such an answer allows better comparison of the model with the expected answer.

This way, we obtained 62 facts and questions about them. This number of facts allowed for quick manual verification and the dataset was deemed satisfactory. We can find out if the model knows the correct answer without providing the text it was trained on. Thus, we do not check the model's quality on the training dataset. Additionally, the prompt aimed to choose facts that are not publicly known.

## Training with QLoRA

For the experiments, we used the library https://github.com/unslothai/unsloth [10], which integrates into frameworks from HuggingFace [9] and allows training many open-source models faster and with less GPU memory, reducing experiment costs.

Since the expert interaction was mimicked using chat dialogs, the model used should support this communication method (chat markup language). Generally, these are instruct versions of the respective models.

The base model chosen was:
- unsloth/mistral-7b-instruct-v0.3-bnb-4bit

This model is already quantized to 4-bit (and optimized) version of mistralai/Mistral-7B-Instruct-v0.3.

This choice is a compromise between the quality provided by the model and the resources required for the experiment. The model unsloth/llama-3-8b-Instruct-bnb-4bit was not used because some errors in the chat message formation mechanism were found at the beginning of the experiment (solved in the latest unsloth releases).

In the LLM training environment, there are different opinions on whether it is possible to memorize facts using PEFT. Most agree that PEFT allows for good adaptation of the response style, but the number of parameters is insufficient to memorize many facts. However, this fact indicates how to fix it. For example, the number of additional parameters in QLoRA is set by the parameter r. Empirical evidence suggests that with r<=16, LLM primarily learns the style, while with r >=64, good fact memorization capabilities are added.

## Some Technical Aspects

As mentioned earlier, we started experiments with a dataset size of 400 KB of text. In this case, we encountered a situation where the model, after training, began to generate repeating sequences up to MAX_LEN tokens. Enriching the dataset with additional fragments of different lengths more or less reliably prevented such behavior. It seems more data was simply needed. The base model did not suffer from infinite generation.

Then it was decided to replenish the dataset because using the same fragments often leads to model overfitting.

Based on the hypothesis that this helped, we have the following hypothesis about the situation:

- Since the persona's communication style was significantly different from how the assistant model usually responds, the model initially tries to adapt to the style. This leads to an increase in loss on each token. As a result, the ability to respond correctly to relatively complex chat markup syntax begins to be lost as it has less impact on the overall loss.
- Only after style adaptation parameters are adjusted does the model start to re-adapt to the markup syntax.
- Therefore, possible solutions to the problem:
    - Provide the model with more diverse data
    - Adapt the loss function to give more weight to markup tokens (we did not experiment with this)

# Quality Assessment

Assessing the quality of LLM performance can be extremely difficult because the tasks it solves are often quite specific, for which there is no dataset. Our task can be reduced to a question-answering task [11], for which there are various datasets, but there is no dataset aimed at measuring the ability to memorize facts about a specific person from interviews (or books or something else).

Another disadvantage of common QA dataset approaches is that numerical metrics such as BLEU, ROUGE, and others are often used for scoring. These metrics pay more attention to the words and phrases used. They may evaluate the style well with some additions, but the fact evaluation in real cases will be significantly lost due to the large number of words and phrases determining the style.

Meanwhile, LLMs themselves can help in such assessment. In our approach, we developed some principles used in the TruLens library [12] for our task. Namely:
- We do not try to calculate a direct metric but compare the answers of different models.
- A LLM compares the answers using the instructions we set.

This approach significantly simplifies the creation of test datasets and allows comparing quite complex substances such as style and facts.

This approach also has limitations - the obtained metrics are comparatively noisy. Besides, pairwise comparisons require more comparisons in an attempt to find the best model. However, they provide fairly confident signals in case of quality changes noticeable to a human.

As mentioned earlier, our ultimate goal is to imitate the expert's work. First and foremost, this is their "perspective." The decisions they make in various situations. Intuitively, this should be at the level of identifying and memorizing facts. And a little bit of communication style (although this is not necessarily). This explains the list of indicators by which we evaluate our model.

For both comparisons, we used GPT-3.5-turbo.

## Style

The style metric is used to assess how much the model's answers correspond to the text style given in the training data. In the context of LLM training, this means that the style metric evaluates how much the model's answers resemble those given by the interviewee in the original data in the following aspects:

- **Lexical similarity**: Comparison of vocabulary and use of specific words and phrases.
- **Syntactic structure**: Evaluation of the similarity of sentence structures, frequency of using complex and simple sentences.
- **Tonality and emotional coloring**: Assessment of how much the model's answers correspond to the emotional tone of the original texts, whether formal, informal, friendly, neutral, etc.
- **Stylistic devices**: Identification of the use of specific stylistic devices such as metaphors, analogies, humor, etc.

To measure this indicator, we started with a fairly simple prompt that satisfied us:

```
I'll give you the real message of some person in the
interview and two fragments (A and B).
Your task is to tell me, which fragment is closer to the
original by style?

[real message]
{original}
[/real message]

[message A]
{message_a}
[/message A]

[message B]
{message_b}
[/message B]

Your answer should contain only one letter of the winner
or sign '=' if both variants are nearly equal. And nothing
else
Examples of the answer:
A

or

B

or

=
```

To understand which fragment better fits the original style, we did the following:

- In the test dataset, we took dialogues similar to those used for training but did not show the real answer of the interviewee to the model. At the same time, the real answer was substituted as {original} in the prompt above.
- When comparing different models, their answers were substituted twice - as {message_a} and as {message_b}. This neutralized the dependence of the answer on the order of occurrence. According to our experiments, this dependence exists. Depending on the prompt details, the model may have a bias either towards the first or second (by order) model.

Why didn't we use the approach practiced in the TruLens library with its Feedback Functions [13]? If we took such an approach, we would collect metrics from each model separately and sort them like regular metric indicators instead of comparing pairwise. And the prompt/request to the model would look something like this:

"Here is a sample response {source} and here is the model's response {message}. Rate how much the model's response resembles the sample in style from 1 to 10."

Unfortunately, practice shows that in this case, when the model has only 2 answers and needs to give a numerical measure in the range of 1 to 10, it usually responds "seven-eight." Inside the team, we even called this the "7/8 problem." Thus, the metric is measured with significant error and can only be applied to large test datasets. It has little to do with the percentages operated by regular metrics.

But if the model can compare fragments directly, it confidently and reasonably chooses the better one.

Even at the preliminary experiment stage, we saw that a volume of 400 KB of plain text (about half of it from the interviewee) was sufficient for the model unsloth/mistral-7b-instruct-v0.3-bnb-4bit with QLoRA adapters r=8…128 to quite confidently start repeating the style after one epoch of training. Further training slightly improved the style match but not significantly. Therefore, we then focused on fact metrics.

## Fact Memorization

Identifying facts present in both the original and generated answers turned out to be a more complex process. The initially written style-metric-like prompt gave unpredictable results that did not correspond to what we saw with the naked eye.

Therefore, the next step was to use the CoT (chain-of-thought) style, where the model had to reason before making a choice. This slightly improved the situation, provided additional information about what was happening, and revealed issues:

- We need to evaluate not only the identified facts but also the ones hallucinated by the evaluating model.
- If the model is allowed to provide the final score, it gets confused about the importance of facts (not always justified) and may give a higher score to the model with fewer fact overlaps based on its reasoning.

After some approach evolution, we developed the following assessment approach:
1. The answer is formed in JSON mode from the OpenAI API, simplifying further parsing.
2. By setting a chain of reasoning, we force the model to sequentially form the following fields:
    2.1. A list of facts found in the original answer.

2.2. Which facts from 2.1 can be found in fragment A.
    2.3. Facts present in fragment A but absent in the original answer.
    2.4. Which facts from 2.1 can be found in fragment B.
    2.5. Facts present in fragment B but absent in the original answer.
 3. Having this answer, we can now calculate TP/FP/FN for each model, based on which we can get numbers corresponding to standard metrics like Precision, Recall, F1, etc. We do this, not the model, based on the structure filled in step 2.
 4. Similarly to the style metric, we repeat the assessment twice, swapping the evaluated fragments.

The prompt for fact assessment turned out to be significantly longer than for style, so we do not provide it in the article but recommend referring to the experiment source code on GitHub.

This approach, after some prompt polishing to mitigate the main format issues, allows obtaining results corresponding to what we observed with the naked eye (for the selected subset of results).

A small note - for fact assessment, it seems we can provide the model-judge only with the original answer and the answer from one model. But practice shows that for the same original fragment, depending on the compared generated fragment, different facts are often found in the original text. Depending on how different the compared fragments are, the model finds facts at different abstraction levels. Therefore, we again settled on the current pairwise comparison scheme of finetuned models. But achieving the possibility of obtaining numerical metrics without comparison would provide additional assessment capabilities: direct sorting of metrics and reducing the effort to find the best model.

During the work on fact assessment, we also realized the types of facts. For our work, we would divide them into 3 types:
 1. **Common knowledge facts**. These facts may be known not only to the finetuned model (or received through RAG) but also to the base one that learned these facts during initial training. To reduce the impact of this type of knowledge on our model, we tried to evaluate facts on the freshest interviews (although they might talk about earlier events).
 2. **Persona-specific facts** that appear in different interviews. In this sense, they are similar to common knowledge facts but still require the model to be finetuned on them.
 3. **Very specific facts** that may be mentioned once. To successfully extract such facts, it is better to use some associative storage with context addition - i.e., the same RAG scheme. Nevertheless, we expect that the model can memorize such facts with a certain level of repetition.

# Experiment Results

## Style

As seen from the style comparison prompt, the result of the comparison is which of the proposed options is closer in style to the original or they are approximately equal. Each comparison for each sample is done twice (changing the order of generated fragments from different models).

In our experiment, we took the test dataset, finetuned the model (parameters used in the article are hardcoded in the training script) on the training dataset, and then compared different models on the test dataset (containing 169 sub-dialogs). We were interested in how much material is needed for the presentation style to resemble the original persona. Preliminary experiments showed that 4-5 hours of interviews (about 400 KB plain-text) allow training the model to respond in a similar style in one pass. But with such volume, it did not train well on chat syntax and often fell into infinite generation. At the same time, we have about a day of interviews, which might be somewhat excessive for understanding how much material is needed. Therefore, we used half the dataset as the "quantum" of material and compared such models:

| Model | Code in tables |
|---|---|
| Base, unsloth/Mistral-7B-Instruct-v0.3 | base |
| Finetuned on 0.5 of the training dataset | 0.5 |
| Finetuned on 1.0 of the training dataset | 1.0 |
| Finetuned on 1.5 of the training dataset | 1.5 (full dataset and supplemented by half) |
| Finetuned on 2.0 of the training dataset | 2.0 (two epochs on the dataset) |

In this experiment, we did not investigate whether the model was trained or overtrained on facts or other characteristics. We only looked at the style of responses.

Assessment was conducted on a test dataset containing 169 sub-dialogs absent from the training set.

Since the evaluation is done using ChatGPT (model gpt-3.5-turbo), even at zero temperature, it can give different results. We checked this point - indeed, repeated runs on the same data gave results differing on average by 1.39% (in terms of the number of test samples for which the model gave different verdicts).

On the one hand, this is quite inconvenient - when using metrics like F1, BLEU, METEOR, we get a result that does not change if we recalculate it again. On the other hand, there is a certain opportunity here. Such uncertainty can be used to avoid overconfidence when assessing the model based on tenths of a percentage metric. This metric automatically provides us with a range of values indicating that changes within these values will not cause significant differences in quality.

Table 1. Raw data (each comparison was done twice):

|      | 0.5 |     |   | 1.0 |     |    | 1.5 |     |   | 2.0 |     |    |
|------|-----|-----|---|-----|-----|----|-----|-----|---|-----|-----|----|
|      | A   | B   | = | A   | B   | =  | A   | B   | = | A   | B   | =  |
| base | 142 | 196 | 0 | 148 | 190 | 0  | 154 | 182 | 2 | 141 | 197 | 0  |
|      | 148 | 190 | 0 | 145 | 193 | 0  | 153 | 185 | 0 | 142 | 196 | 0  |
|      |     |     |   |     |     |    |     |     |   |     |     |    |
| 0.5  |     |     |   | 144 | 176 | 18 | 161 | 172 | 5 | 147 | 185 | 6  |
|      |     |     |   | 139 | 182 | 17 | 158 | 176 | 4 | 143 | 188 | 7  |
|      |     |     |   |     |     |    |     |     |   |     |     |    |
| 1.0  |     |     |   |     |     |    | 165 | 167 | 6 | 163 | 164 | 11 |
|      |     |     |   |     |     |    | 172 | 159 | 7 | 169 | 160 | 9  |
|      |     |     |   |     |     |    |     |     |   |     |     |    |
| 1.5  |     |     |   |     |     |    |     |     |   | 159 | 175 | 4  |
|      |     |     |   |     |     |    |     |     |   | 162 | 169 | 7  |

In this table, column A shows how many times the answer from model A (first column) was better, B shows how many times the answer from model B (first row) was better, and the = symbol indicates how many times the model decided that the answers were approximately equal in style. Each comparison involves 169 samples compared twice.

Table 2. Differences in two attempts on the same data:

|      | 0.5 |   |   | 1.0 |   |   | 1.5 |   |   | 2.0 |   |   |
|------|-----|---|---|-----|---|---|-----|---|---|-----|---|---|
|      | A   | B | = | A   | B | = | A   | B | = | A   | B | = |
| base | 6   | 6 | 0 | 3   | 3 | 0 | 1   | 3 | 2 | 1   | 1 | 0 |
| 0.5  |     |   |   | 5   | 6 | 1 | 3   | 4 | 1 | 4   | 3 | 1 |
| 1.0  |     |   |   |     |   |   | 7   | 8 | 1 | 6   | 4 | 2 |
| 1.5  |     |   |   |     |   |   |     |   |   | 3   | 6 | 3 |

Thus, summing up the data from tables 1 and 2, we get:

- Total number of answers: 6760
- Sum of differences in repeated experiments: 94
- This means the average relative difference in repeated experiments is 1.39%

Assuming the differences follow a normal distribution, if we want to be 99% confident that one model is better than another, the difference between A and B values must exceed:

- 1.39% * z(0.99) = 1.39% * 2.576 = 3.58% (from the number of samples)
- If we want to be 95% confident (z(0.95)=1.96), the difference should exceed 2.72%

For 2*169=338 samples from our test dataset, if the differences between A and B are less than 6.62 (rounded to 7), we cannot be 95% or more confident that the dataset is better. We can try to increase the accuracy of the assessment with repeated runs and averaging. Another possible option could be refining the evaluation prompt to make the assessment more stable, reducing the confidence interval.

Table 3. Pairwise comparisons for style characteristic with percentage difference:

|      | 0.5 | | | 1.0 | | | 1.5 | | | 2.0 | | |
|------|-----|-----|-------|-----|-----|-------|-----|-----|-------|-----|-----|-------|
|      | A   | B   | WinB  | A   | B   | WinB  | A   | B   | WinB  | A   | B   | WinB  |
| base | 290 | 386 | 96    | 293 | 383 | 90    | 307 | 367 | 60    | 283 | 393 | 110   |
| %    |     |     | 14.20%|     |     | 13.31%|     |     | 8.88% |     |     | 16.27%|
| 0.5  |     |     |       | 283 | 358 | 75    | 319 | 348 | 29    | 290 | 373 | 83    |
| %    |     |     |       |     |     | 11.09%|     |     | 4.29% |     |     | 12.28%|
| 1.0  |     |     |       |     |     |       | 337 | 326 | -11   | 332 | 324 | -8    |
| %    |     |     |       |     |     |       |     |     | -1.63%|     |     | -1.18%|
| 1.5  |     |     |       |     |     |       |     |     |       | 321 | 344 | 23    |
| %    |     |     |       |     |     |       |     |     |       |     |     | 3.40% |

Based on pairwise comparisons, we can say that:

- All finetuned models are statistically significantly better than the base model.
- All models trained on the full dataset and more are statistically significantly better than the model trained on half the dataset.
- The model trained on one epoch on the full dataset (1.0) is insignificantly better than the models trained more (1.5, 2.0).

Of course, the obtained values depend on the base model, training hyperparameters, the diversity of the persona's personal style, but we can already say that:

- About 1 MB of interviews or similar materials (about 500 KB of persona text, up to 12 hours of audio dialogs) is sufficient to achieve similarity in presentation style confirmed by visual checks.
- 2 MB of text is generally sufficient - for further quality improvement, it's better to think about tuning training hyperparameters.
- Likely, significantly less text would be enough for good style imitation with several epochs of training, but it is necessary to control other target training parameters and the possibility of overfitting (however, using QLoRA significantly reduces its risk).

## Fact Memorization

In this experiment, we were interested in how much finetuning (using QLoRA) makes the model memorize certain facts. Therefore, the standard version with a training and test dataset did not fit - the test dataset simply wouldn't contain single facts from the training one and vice versa. Thus, we exclude facts related to the third group and can only check facts from the first and second groups (common knowledge and frequently mentioned in the personality's interviews). In this part of the experiment, we proceeded differently:

1. Separated 3 interviews from 2023 (as the most recent) into the **new** group.
2. Marked the remaining interviews as the **old**.
3. Using ChatGPT-4o, extracted 62 facts from the new interviews and prepared the dataset as described in the Fact Q/A dataset section.

Thus, the test included facts that some models were trained on, but the model could not answer verbatim. To answer correctly, it had to memorize and then correctly insert the fact information.

The following models participated in the tests:

| Model | Code in data |
| --- | --- |
| Base, unsloth/Mistral-7B-Instruct-v0.3 | base |
| Finetuned on the old part of the dataset | old |
| Finetuned on the new part of the dataset and then on the old part of the dataset | new_old |
| Finetuned on the old part of the dataset and then on the new part of the dataset | old_new |

Finetuning involved 2 epochs of training on each part.

The intuition was as follows:

If the model can memorize single data and there is an advantage of new data over previous ones, the models will rank in the following order (from worst to best):

1. Base
2. Trained on "old" interviews - it will not know the facts from the test questionnaire that are rare and single, but can answer those that appear in one form or another in earlier interviews, as well as general knowledge.
3. Trained first on the "new" part of the dataset and then on the "old" part.
4. Trained first on the "old" part of the dataset and then on the "new."

|  | old | | | | | | new_old | | | | | | old_new | | | | | |
|---|---|---|---|---|---|---|---|---|---|---|---|---|---|---|---|---|---|---|
|  | TP | FP | FN | Pr | Rec | F1 | TP | FP | FN | Pr | Rec | F1 | TP | FP | FN | Pr | Rec | F1 |
|  | 98 | 283 | 98 | 25.7% | 50.0% | 33.9% | 100 | 314 | 103 | 24.1% | 49.2% | 32.4% | 97 | 285 | 106 | 25.3% | 47.7% | 33.1% |
| base | 96 | 548 | 100 | 14.9% | 48.9% | 22.8% | 107 | 548 | 96 | 16.3% | 52.7% | 24.9% | 99 | 534 | 104 | 15.6% | 48.7% | 23.6% |
|  |  |  |  |  |  |  |  |  |  |  |  |  |  |  |  |  |  |  |
|  |  |  |  |  |  |  | 110 | 261 | 94 | 29.6% | 53.9% | 38.2% | 109 | 233 | 98 | 31.8% | 52.6% | 39.7% |
| old |  |  |  |  |  |  | 102 | 243 | 102 | 29.5% | 50.0% | 37.1% | 114 | 266 | 93 | 30.0% | 55.0% | 38.8% |
|  |  |  |  |  |  |  |  |  |  |  |  |  |  |  |  |  |  |  |
|  |  |  |  |  |  |  |  |  |  |  |  |  | 106 | 246 | 101 | 30.1% | 51.2% | 37.9% |
| new_old |  |  |  |  |  |  |  |  |  |  |  |  | 105 | 260 | 102 | 28.7% | 50.7% | 36.7% |

Table 4. Pairwise comparisons of generation results from different models:
We use the F1 score for quality comparison. Note that we have a pseudo-F1 valid only within the comparison of two models and cannot be directly used to compare models from different "tournaments."

Table 5. Which model is better and by how much:

|  | old | | new_old | | old_new | |
|---|---|---|---|---|---|---|
|  | F1 diff | Winner | F1 diff | Winner | F1 diff | Winner |
| base | 11.11% | old | 7.47% | new_old | 9.48% | old_new |
| old |  |  | 1.10% | new_old | 0.87% | old_new |
| new_old |  |  |  |  | 1.21% | old_new |

Here we can conclude that:

- All finetuned models performed statistically significantly better than the base model on the fact questionnaire. This means they could memorize at least second-type facts.

- Models with added interviews containing facts from the questionnaire performed better than those trained only on old data, but these differences are within the margin of error, requiring further experiments to prove the improvement.
- Direct comparison showed that the model finetuned first on old interviews and then on new ones was slightly better than in the reverse order, but this advantage is also small, requiring confirmation in additional studies.

In cases where the effect was at the statistical significance boundary, visual checks usually showed that it was difficult for a human to unequivocally choose a winner, indirectly indicating the correctness of the assessment methodology.

An important secondary criterion of the applied assessment methodologies (both style and factuality) is their transitivity (considering the small noise associated with the stochastic nature of LLM generation). If pairwise comparisons reveal model_a > model_b, model_b > model_c, then model_a > model_b (where ">" means the left model received a higher quality score than the right one). This is important because otherwise, identifying the best model would be impossible.

# Conclusions

- To train a model to the style of a person, 4-5 hours of their interviews may be sufficient.
- In our experiments, QLoRA allows the model to memorize facts, but mainly of the second type - those repeated in various forms during communication with the persona. For memorizing rare single facts, it's better to use RAG.
- Direct comparison of two models' responses allows measuring the quality of various features that are difficult to reduce to standard numerical metrics. Thus, we can well compare the style and factual content of the answers separately.